\documentclass[11pt]{article}
\usepackage[final]{acl}   
\usepackage{times}
\usepackage{latexsym}
\usepackage[T1]{fontenc}
\usepackage[utf8]{inputenc}
\usepackage{microtype}
\usepackage{graphicx}
\usepackage{subcaption}
\usepackage{booktabs}
\usepackage{array}
\usepackage{enumitem}
\usepackage{amssymb}
\usepackage{xcolor}
\usepackage{url}

\newcommand{\yes}{\textcolor{green!55!black}{\checkmark}}
\newcommand{\pmark}{\textcolor{orange!85!black}{$\sim$}}
\newcommand{\no}{\textcolor{red!70!black}{$\times$}}

\title{PolyInterview: An LLM-based Platform for Immersive Mock Interview\\ Practice with Comprehensive Multimodal Assessment}

\author{
  \bfseries Zhiyuan Wen$^{1}$, Jiannong Cao$^{1}$, Kelly Chan$^{1}$, Zijian Wang$^{1}$, Chen Chen$^{1}$, Xiaoyun Liu$^{1}$, Jianing Yin$^{1}$, Zhuo Li$^{2}$ \\[3pt]
  \normalfont $^{1}$Department of Computing, The Hong Kong Polytechnic University, Hong Kong, China \\
  \normalfont $^{2}$School of Artificial Intelligence, Chongqing University of Posts and Telecommunications, China \\[3pt]
  \normalfont\texttt{\{zyuanwen, jiannong.cao, kelly.chan, zi-jian.wang\}@polyu.edu.hk} \\
  \normalfont\texttt{\{chen03.chen, xiaoyun.liu, jianing-laetitia.yin\}@connect.polyu.hk} \\
  \normalfont\texttt{cliff.zhuo.li@gmail.com}
}

\begin{document}
\maketitle

\begin{abstract}
Preparing for job interviews is important for securing desired positions, yet realistic practice remains difficult to access: real interviews are infrequent, expert mock coaching is costly, and self-practice offers neither adaptive dialogue nor structured assessment. Existing systems typically address only parts of this need through fixed question sequences, limited communication channels, or feedback with little supporting evidence. We present \textbf{PolyInterview}, an LLM-based platform for immersive mock interview practice with comprehensive multimodal assessment. PolyInterview uses the target job description and CV to generate questions tailored to the role and candidate, conducts multi-turn spoken interviews with a lip-synced digital human interviewer that asks answer-aware follow-up questions, and evaluates response content, vocal delivery, and non-verbal behavior. Four parallel evaluators produce 13 behavior-level features that are aggregated into 10 assessment aspects and two competency tracks. Guided by the KSA and STAR frameworks, the report links each score to behavioral evidence and actionable recommendations. PolyInterview is publicly accessible\footnote{PolyInterview is deployed for public access. The demo video and system link are available \href{https://dannywang1922.github.io/polyinterview}{here}.}. Its current all-account snapshot contains 101 accounts, 1{,}564 interview sessions, 7{,}665 generated questions, and 1{,}422 five-stage question sets. Generated questions are more closely aligned with their matched job description than with cross-role job descriptions in 93.7\% of sessions. An evaluation by ten experts found strong question plans and actionable feedback.
\end{abstract}

\begin{figure}[t]
  \centering
  \includegraphics[width=0.9\linewidth]{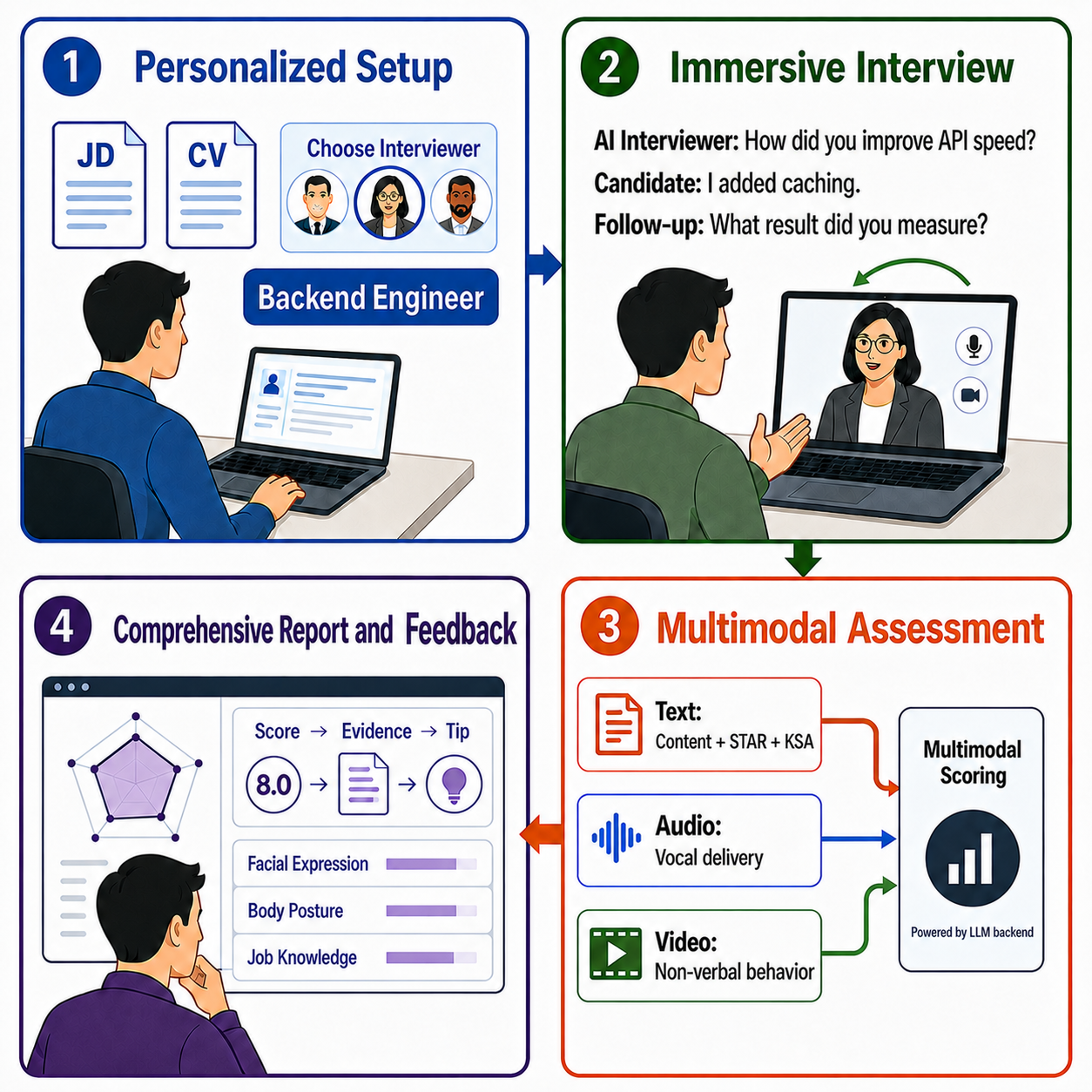}
\caption{PolyInterview's workflow, from personalized setup and immersive interviewing to multimodal assessment and comprehensive reporting and feedback.}
  \label{fig:teaser}
\end{figure}

\section{Introduction}
\label{sec:intro}
Effective preparation for job interviews is critical to securing a desired position. During an interview, both candidates' role-relevant knowledge and skills and their ability to communicate experience and reasoning clearly are evaluated. Despite the importance of interview preparation, realistic practice remains difficult to access: real interviews are infrequent, expert mock coaching is costly, and self-practice offers neither adaptive dialogue nor structured assessment. These constraints motivate an accessible platform that can simulate multi-turn interview interactions and provide comprehensive assessment and feedback.

Commercial products and research prototypes have expanded access to interview practice, yet, as summarized in Table~\ref{tab:compare}, they do not jointly support role-conditioned multi-stage questioning, answer-aware probing, and theory-grounded assessment of textual, vocal, and non-verbal behavior within an integrated practice environment. For commercial tools, products such as Google Interview Warmup, Yoodli, and Big Interview support question rehearsal or automated feedback, but generally rely on fixed question sequences, assess only selected communication channels, or provide limited transparency into how feedback is derived. For research prototypes, virtual interview agents have explored conversational realism \citep{hoque2013mach,anderson2013tardis,smith2014virtual}; LLM-based mock interviewers support flexible question generation \citep{li2023ezinterviewer,sun2025mockllm,daryanto2025conversate}; and automated video interview systems analyze non-verbal behavior \citep{naim2018automated,hemamou2019hirenet}.

To address these limitations, we present \textbf{PolyInterview}, an LLM-based platform for immersive mock interview practice with comprehensive multimodal assessment (Figure~\ref{fig:teaser}). PolyInterview personalizes each mock interview by using the target job description (JD) and the candidate's CV to generate tailored questions. A lip-synced digital human interviewer creates an immersive interview setting, while LLM-based agents sustain multi-turn interaction and generate answer-aware follow-up questions. For comprehensive assessment, we combine LLM-based analysis of response content, speech analysis, and vision-language model (VLM)-based analysis of non-verbal behavior. Guided by the Knowledge, Skills, and Abilities (KSA) framework \citep{peterson2001onet} and the Situation--Task--Action--Result (STAR) structure for organizing evidence in patterned behavior description interviews \citep{janz1982behavior}, the assessment captures two complementary dimensions of interview performance: the extent to which candidates demonstrate knowledge, skills, and experience relevant to the target role, and how clearly and effectively they communicate them. The resulting 13 behavior-level features are aggregated into 10 assessment aspects and two competency tracks, with each score linked to behavioral evidence and actionable feedback.

PolyInterview's current all-account snapshot contains 101 accounts and 1{,}564 interview sessions. It includes 7{,}665 generated questions spanning 83 position titles, with 1{,}422 session question sets covering the complete five-stage interview structure. A lexical alignment analysis further shows that generated question sets are more closely aligned with their matched JD than with cross-role JDs in 93.7\% of sessions, and the matched JD ranks first in 82.4\% of sessions. An evaluation by ten human experts further identified strong question plan quality and actionable recommendations, while revealing follow-up diagnostic depth and response faithfulness as areas for improvement. In summary, our contributions are threefold:
\begin{enumerate}[noitemsep,topsep=2pt,leftmargin=*]
  \item \textbf{Immersive, personalized interview simulation.} We develop an end-to-end workflow that constructs a multi-stage interview from the JD and CV, adapts follow-up questions to candidate responses, and delivers the interaction through a configurable, lip-synced digital human interviewer (\S\ref{sec:journey}--\S\ref{sec:match}).
  \item \textbf{Comprehensive multimodal assessment.} We introduce a pipeline grounded in KSA and STAR that analyzes textual, vocal, and visual behavior through four evaluators, maps 13 features to 10 aspects and two competency tracks, and preserves a traceable path from each score to actionable feedback (\S\ref{sec:scoring}).
  \item \textbf{Deployment and expert evaluation.} We characterize an all-account snapshot of 101 accounts and 1{,}564 sessions, analyze workflow coverage and role alignment, and evaluate question plans, follow-ups, and feedback reports with ten human experts (\S\ref{sec:deploy} and \S\ref{sec:human-expert}).
\end{enumerate}

\begin{table*}[t]
\centering\small
\setlength{\tabcolsep}{4pt}
\begin{tabular}{lccccc}
\toprule
& \textbf{Adaptive} & \textbf{Theory-grounded} & \textbf{Multimodal} & \textbf{Multi-stage} & \textbf{Explainable} \\
& \textbf{follow-ups} & \textbf{assessment} & \textbf{(text/audio/video)} & \textbf{interview} & \textbf{\& traceable} \\
\midrule
\multicolumn{6}{l}{\emph{Commercial tools}} \\
Google Interview Warmup & \no & \pmark & \pmark & \no & \no \\
Yoodli & \no & \pmark & \pmark & \no & \no \\
Big Interview & \pmark & \pmark & \no & \no & \no \\
Final Round AI & \pmark & \pmark & \pmark & \no & \pmark \\
HireVue & \pmark & \pmark & \no\,(removed 2020) & \pmark & \no \\
Interviewing.io (human) & \yes & \yes & N/A & \yes & \pmark \\
\midrule
\multicolumn{6}{l}{\emph{Research systems}} \\
MACH \citep{hoque2013mach} & \no & \no & \yes & \no & \yes \\
TARDIS \citep{anderson2013tardis} & \no & \no & \pmark & \pmark & \pmark \\
VR-JIT \citep{smith2014virtual} & \pmark & \pmark & \no & \pmark & \yes \\
ERICA interviewer \citep{inoue2020jobinterviewer} & \yes & \no & \no & \no & \no \\
EZInterviewer \citep{li2023ezinterviewer} & \pmark & \no & \no & \no & \no \\
MockLLM \citep{sun2025mockllm} & \yes & \no & \no & \no & \no \\
Conversate \citep{daryanto2025conversate} & \yes & \pmark & \no & \no & \yes \\
\midrule
\textbf{PolyInterview (ours)} & \yes & \yes\,(KSA$\times$STAR) & \yes & \yes & \yes \\
\bottomrule
\end{tabular}
\caption{Comparison with representative commercial and research interview systems (\yes\ full, \pmark\ partial, \no\ absent).}
\label{tab:compare}
\end{table*}

\section{System Overview and Design}
\label{sec:system}

\begin{figure*}[t]
  \centering
  \includegraphics[width=0.9\textwidth]{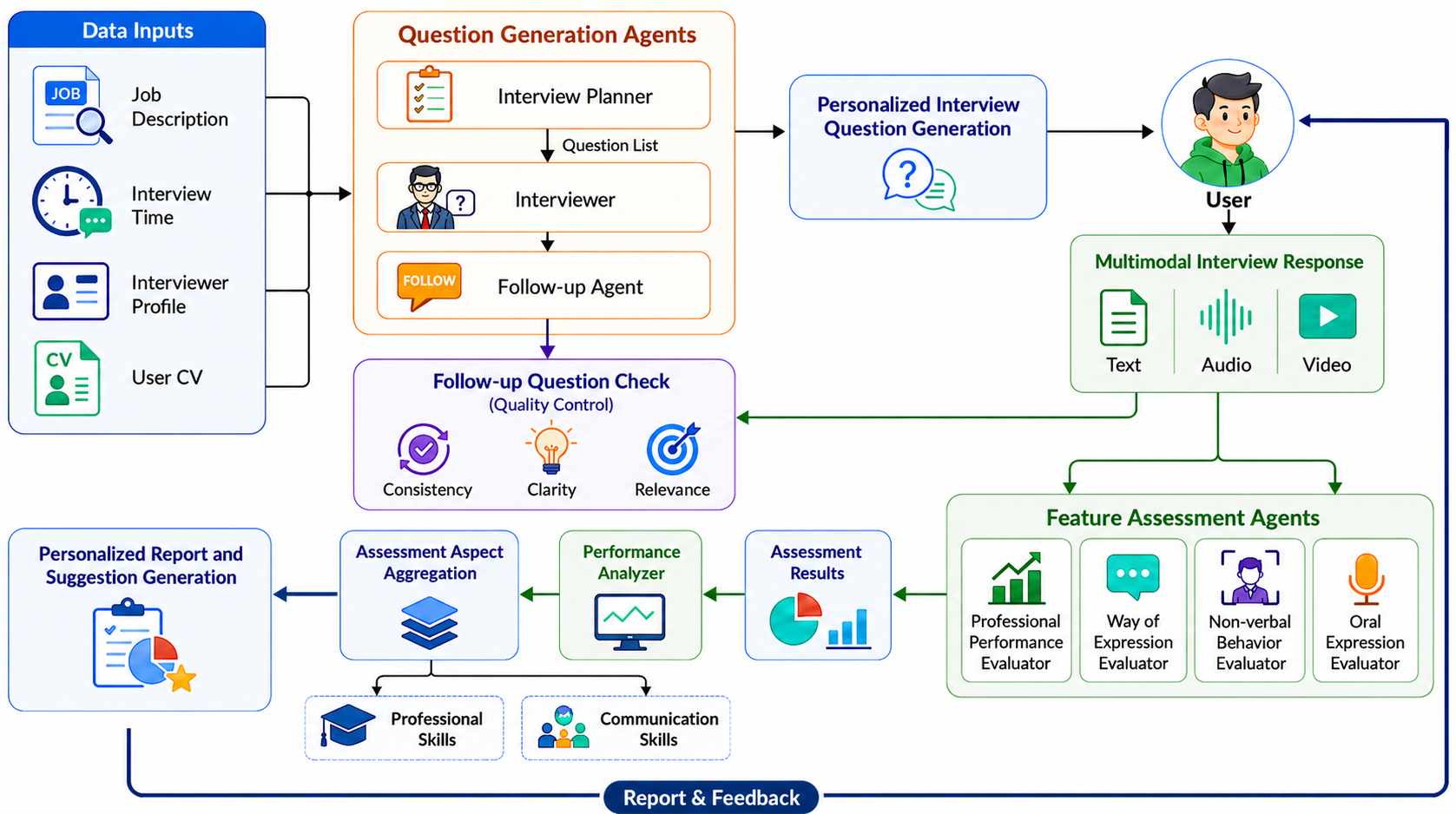}
\caption{PolyInterview's pipeline for personalized questioning, answer-aware follow-ups, multimodal assessment, and comprehensive feedback.}
  \label{fig:pipeline}
\end{figure*}

A PolyInterview session comprises four stages (Figure~\ref{fig:pipeline}): \emph{personalized setup}, \emph{immersive interview}, \emph{multimodal assessment}, and \emph{comprehensive report and feedback}. The candidate configures a personalized session, interacts with the digital human interviewer through multi-turn spoken dialogue, and reviews both per-question and whole-interview feedback. We first describe the interface and user workflow (\S\ref{sec:journey}), followed by the mechanisms for adaptive interviewing (\S\ref{sec:match}) and comprehensive multimodal assessment (\S\ref{sec:scoring}).

\subsection{User Interface and Workflow}
\label{sec:journey}

\begin{figure*}[!t]
  \centering
  \begin{minipage}{0.92\textwidth}
  \centering
  \begin{subfigure}[t]{0.36\linewidth}
    \centering
    \includegraphics[width=\linewidth]{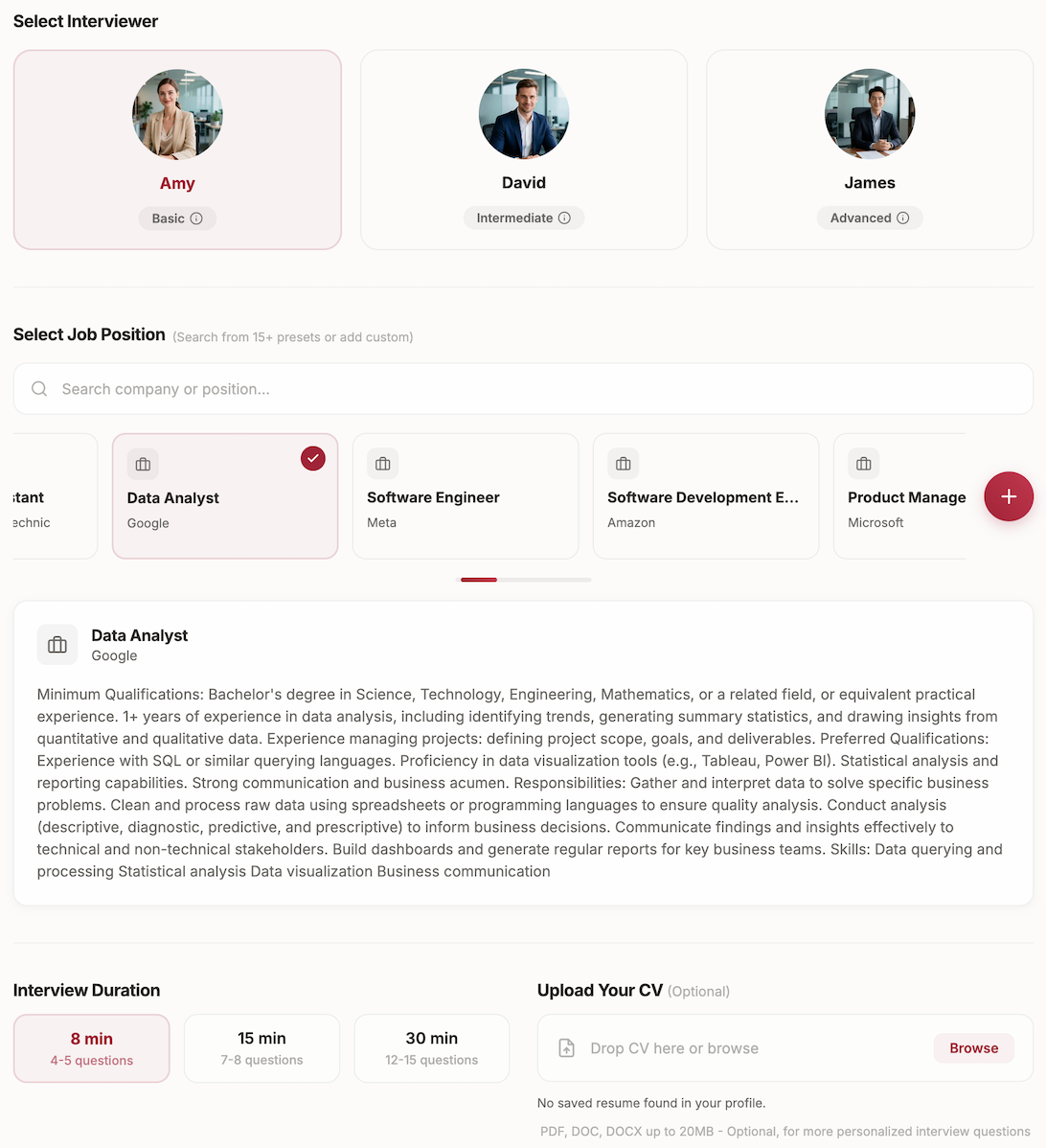}
\caption{Personalized setup.}
    \label{fig:ui-setup}
  \end{subfigure}\hfill
  \begin{subfigure}[t]{0.62\linewidth}
    \centering
    \includegraphics[width=\linewidth]{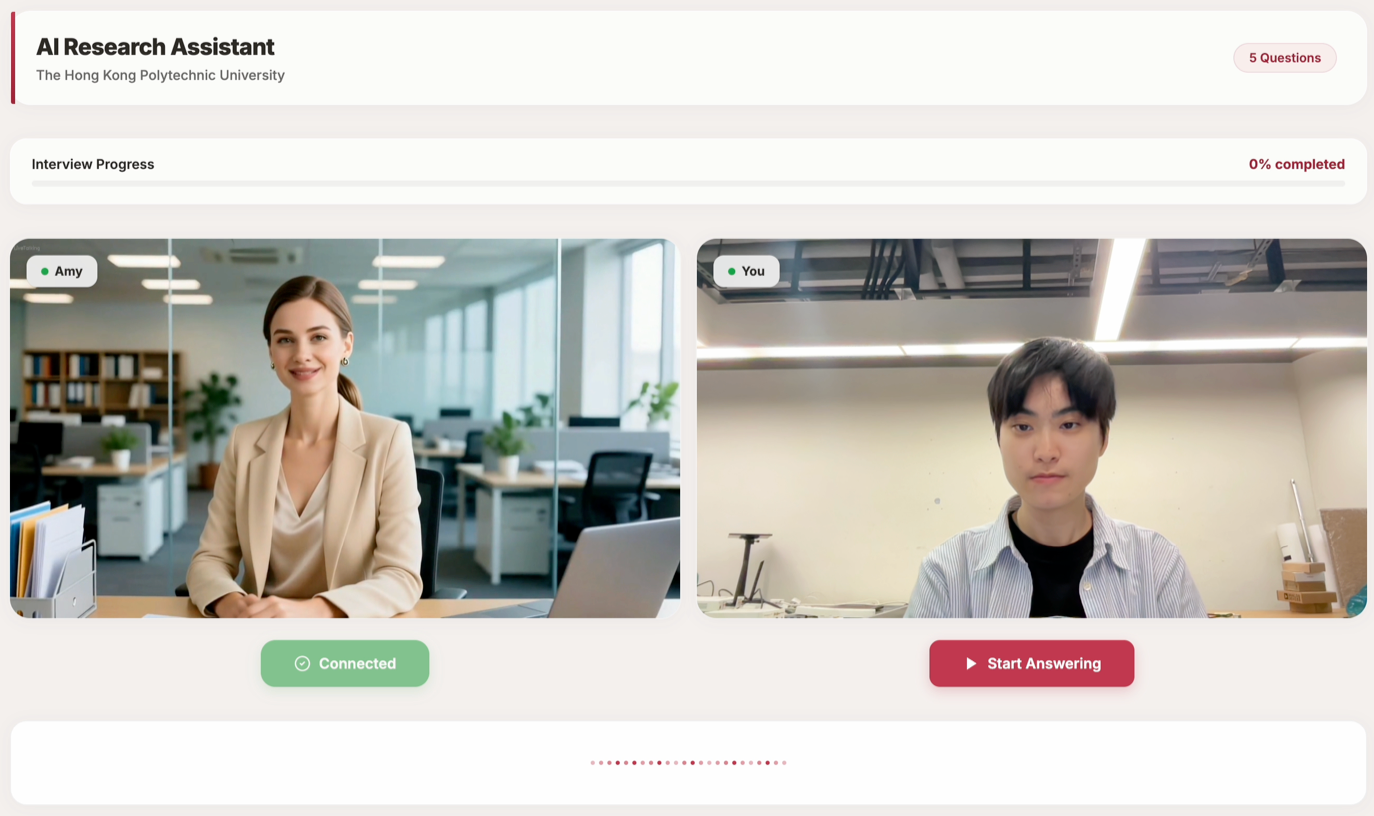}
\caption{Immersive digital human interview.}
    \label{fig:ui-live}
  \end{subfigure}

  \vspace{2pt}
  \begin{subfigure}[t]{0.62\linewidth}
    \centering
    \includegraphics[width=\linewidth]{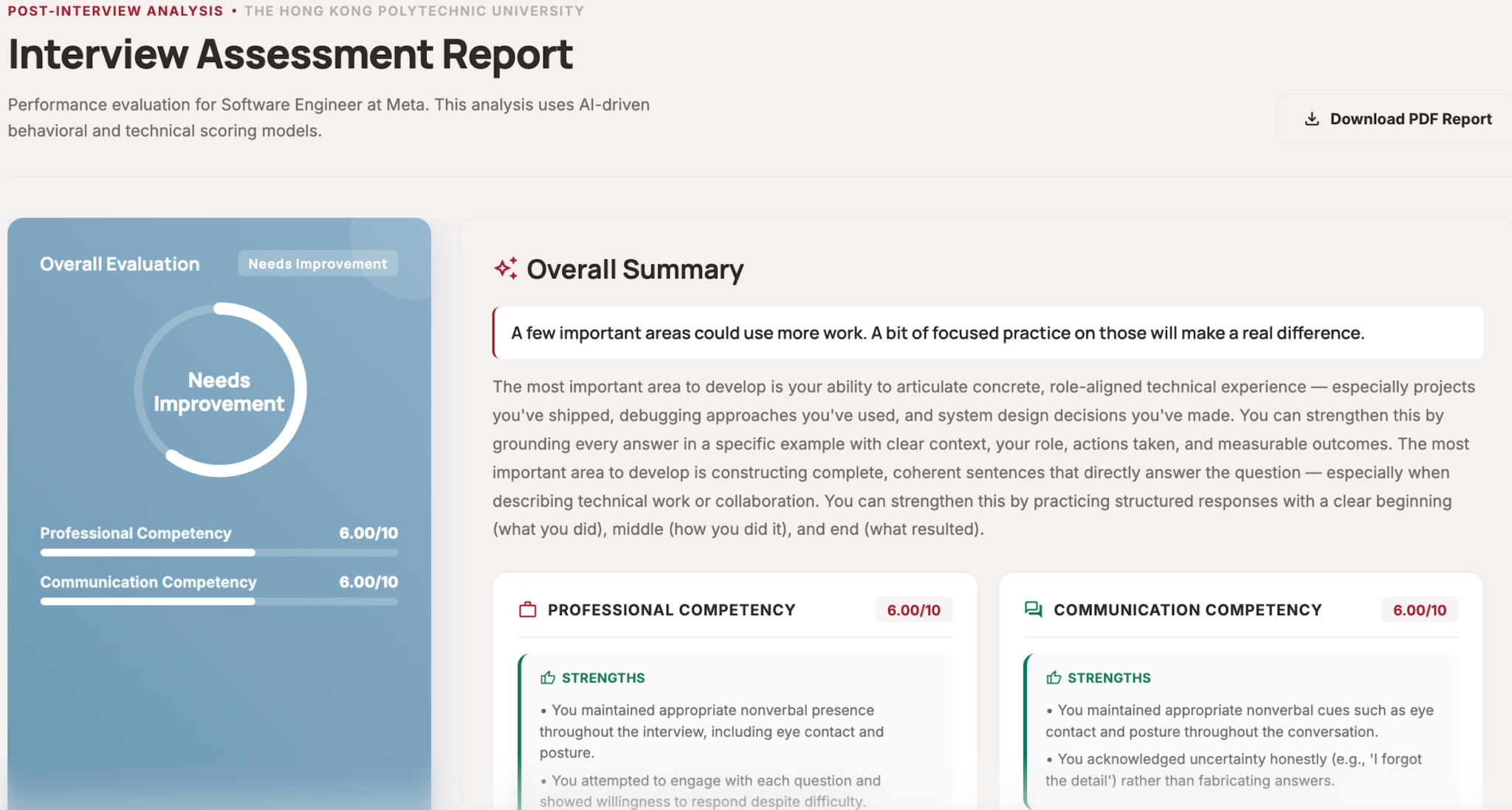}
\caption{Overall assessment summary.}
    \label{fig:report-overall}
  \end{subfigure}\hfill
  \begin{subfigure}[t]{0.36\linewidth}
    \centering
    \includegraphics[width=\linewidth]{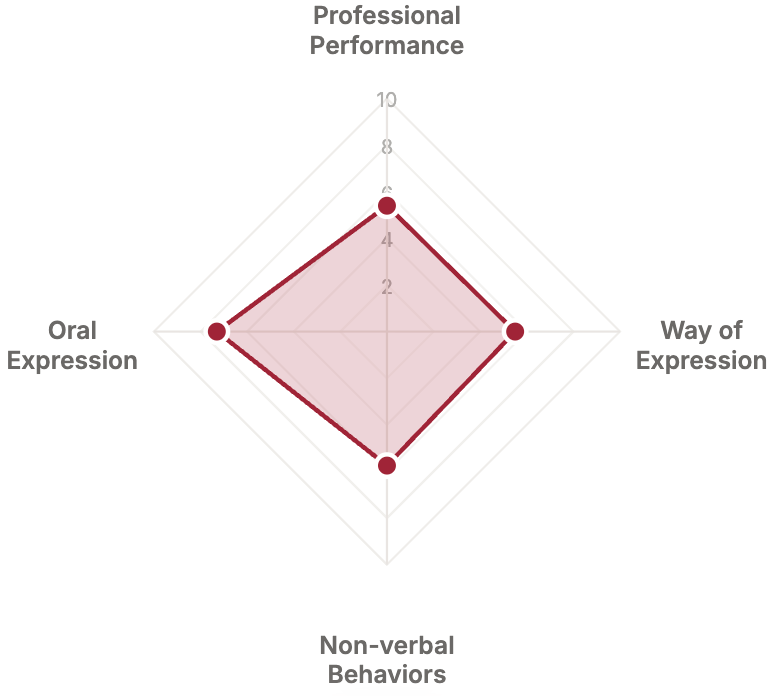}
\caption{Behavior-level feature profile.}
    \label{fig:report-skills}
  \end{subfigure}

  \vspace{2pt}
  \begin{subfigure}[t]{0.38\linewidth}
    \vspace{0pt}
    \centering
    \includegraphics[width=\linewidth]{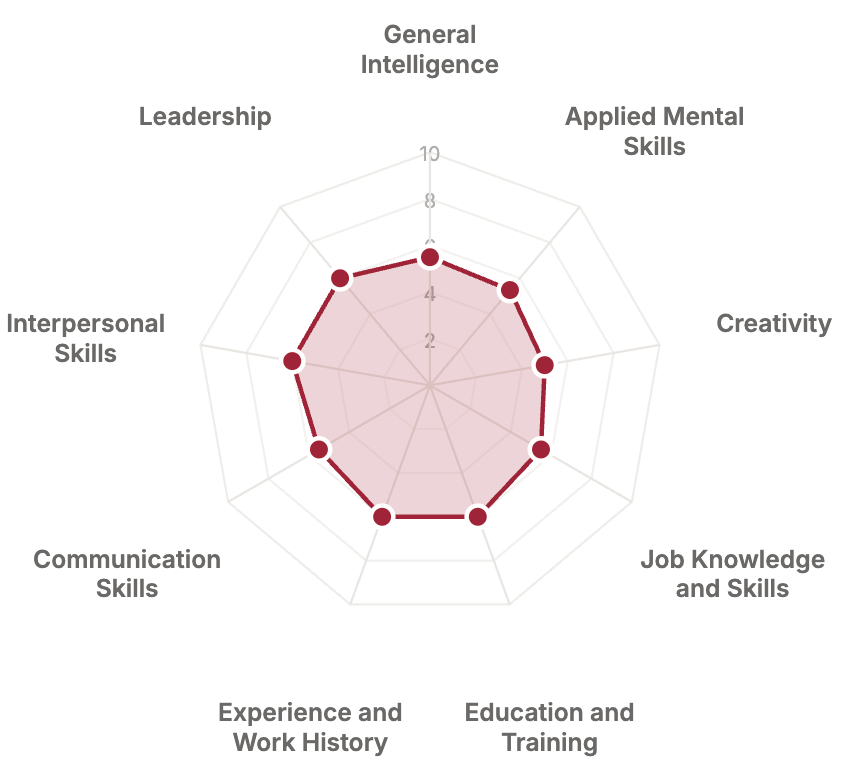}
\caption{KSA-aligned assessment aspect profile.}
    \label{fig:report-radar}
  \end{subfigure}\hfill
  \begin{subfigure}[t]{0.60\linewidth}
    \vspace{0pt}
    \centering
    \includegraphics[width=\linewidth,trim=0 834bp 0 0,clip]{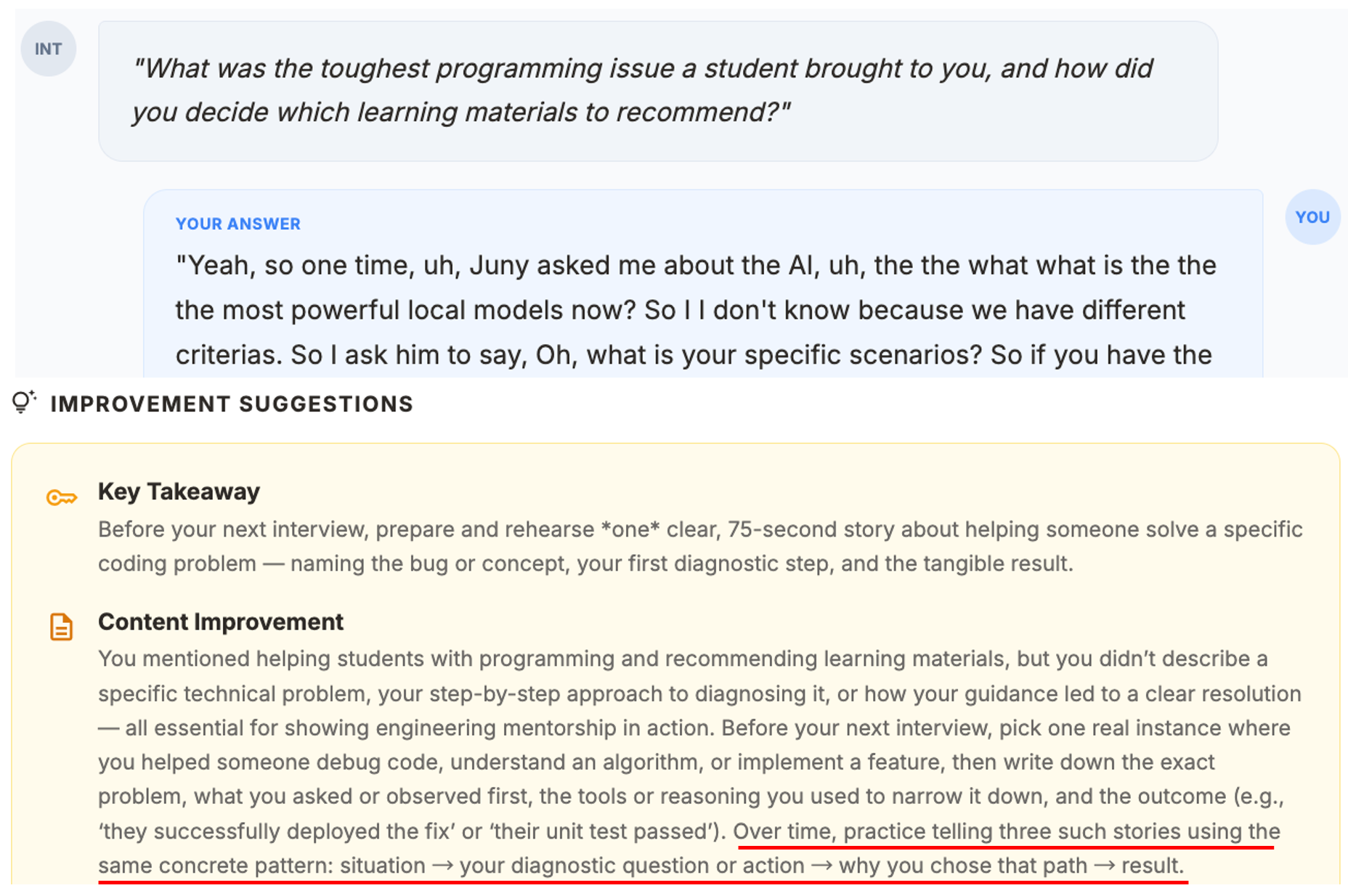}
    \vspace{-1pt}
    \includegraphics[width=\linewidth,trim=0 0 0 610bp,clip]{figures/ui/feedback_q.png}
\caption{Per-question diagnosis and improvement suggestions.}
    \label{fig:ui-feedback}
  \end{subfigure}
  \end{minipage}
\caption{PolyInterview's interface for personalized setup, immersive interviewing, and multilevel assessment results.}
  \label{fig:journey}
\end{figure*}

\paragraph{Interview setup.} The setup interface consolidates the inputs required for personalization on one screen (Figure~\ref{fig:ui-setup}). The candidate selects one of three interviewer personas, specifies a target company, position, and JD, uploads a CV in PDF format, and chooses a session of 8, 15, or 30 minutes. These inputs condition both the content and length of the interview. The JD and CV provide the evidence used to tailor the interview to the role and candidate, while the selected duration determines the question budget.

\paragraph{Immersive live interview.} A lip-synced digital human interviewer presents each question aloud, and streaming speech recognition supports spoken responses. The interface places the interviewer and candidate video side by side and displays connection status, session progress, and response controls (Figure~\ref{fig:ui-live}). For each question, the system records the response transcript, audio, and video, enabling subsequent assessment of response content, vocal delivery, and non-verbal behavior.

\paragraph{Comprehensive assessment results.} After the interview, the interface presents assessment results and feedback from summary to detail. The overall report presents the overall assessment, two competency-track scores, strengths, and improvement priorities (Figure~\ref{fig:report-overall}). Candidates can then inspect the 10 KSA-aligned assessment aspects (Figure~\ref{fig:report-radar}) and drill down to the 13 behavior-level features produced by the four evaluators (Figure~\ref{fig:report-skills}). Finally, the per-question view pairs each prompt and response with an answer-specific diagnosis and improvement suggestions, including STAR-guided phrasing that turns the candidate's experience into a structured account (Figure~\ref{fig:ui-feedback}). This hierarchy connects the assessment framework in \S\ref{sec:scoring} to concrete practice guidance. The complete report can also be exported as a PDF.

\subsection{Adaptive Interviewing}
\label{sec:match}

Personalization operates both before and during the interview: the Interview Planner Agent combines role requirements from the target JD with relevant skills, projects, and experience from the candidate's CV to generate questions tailored to the role and candidate, while the Interviewer Agent produces bounded, answer-aware follow-ups. Appendix~\ref{app:adaptive} provides the detailed question-planning and follow-up workflows.

\subsection{Comprehensive Multimodal Assessment}
\label{sec:scoring}

\begin{figure}[t]
  \centering
  \includegraphics[width=\linewidth]{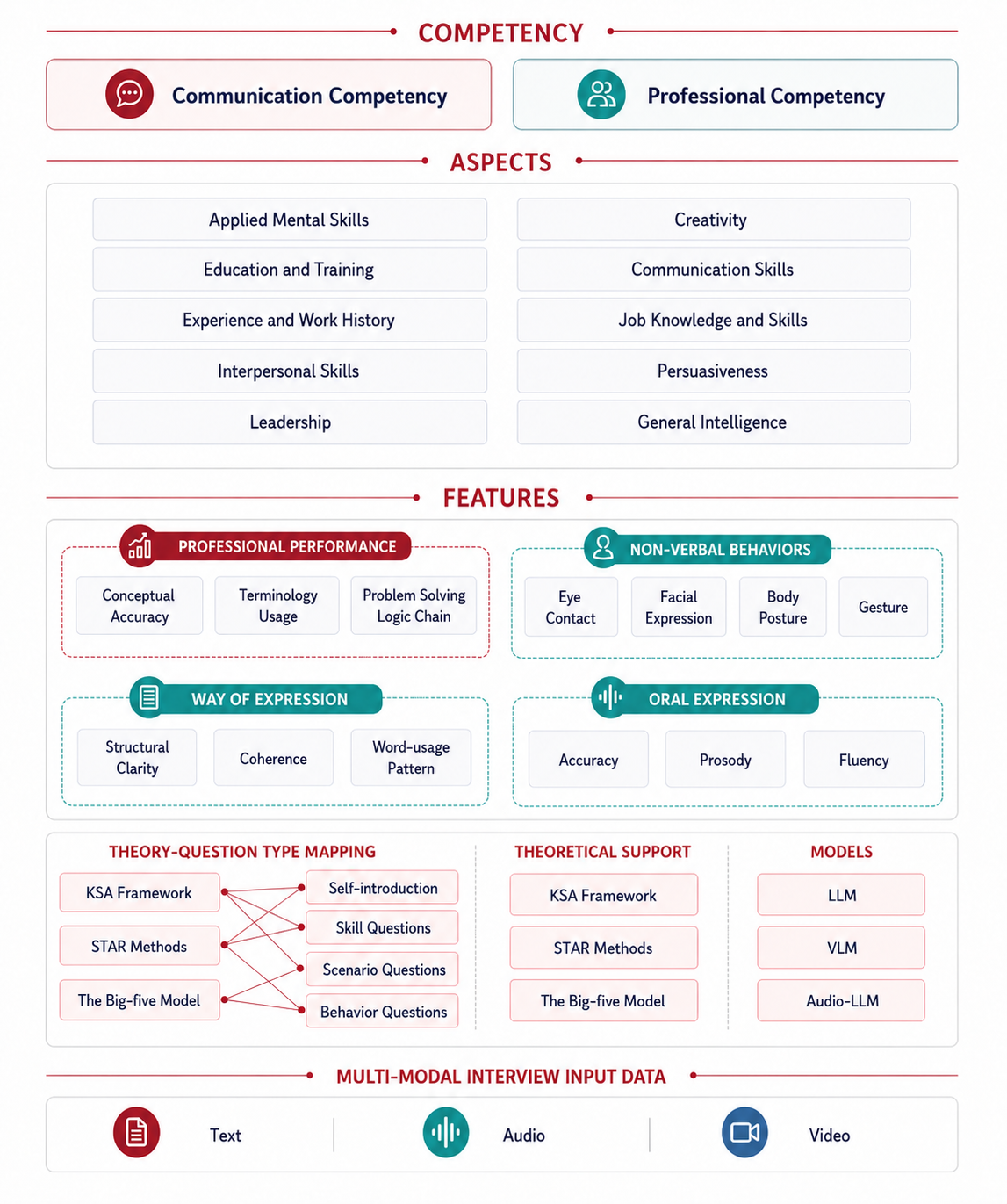}
\caption{Three-layer multimodal assessment from 13 behavior-level features to 10 aspects and two competency tracks.}
  \label{fig:arch}
\end{figure}

Figure~\ref{fig:arch} presents the three-layer assessment architecture. Four evaluators analyze each response in parallel, and their outputs are progressively aggregated from behavior-level features to assessment aspects and competency tracks.

\paragraph{Layer 1: Behavior-level features.} Two LLM-based text evaluators analyze response content and expression. The \emph{Professional Performance} evaluator assesses conceptual accuracy, terminology use, and problem-solving logic, while the \emph{Way of Expression} evaluator assesses structural clarity, coherence, and word usage. A VLM-based \emph{Non-verbal Behavior} evaluator analyzes eye contact, facial expression, body posture, and gesture from the recorded video. A speech-based \emph{Oral Expression} evaluator analyzes pronunciation accuracy, prosody, and fluency. Together, these evaluators produce 13 feature scores on a scale from 0 to 10.

\paragraph{Layer 2: Assessment aspects.} An aggregation agent maps the 13 features to 10 KSA-aligned aspects in three families: \emph{cognitive} (general intelligence, applied mental skills, and creativity), \emph{background} (job knowledge and skills, education and training, and experience and work history), and \emph{social} (communication, interpersonal skills, leadership, and persuasiveness). The social family is informed by the Big Five \citep{mccrae1987validation}, without inferring or reporting personality traits. Each aspect combines primary and secondary features with a 70/30 weighting. Question-category gating restricts assessment to aspects that the current question can reasonably elicit. For example, behavioral questions activate interpersonal skills, applied mental skills, and leadership, whereas skill-QA questions activate job knowledge, general intelligence, and education. Appendix~\ref{app:mapping} details the mapping and gating rules.

\paragraph{Layer 3: Competency tracks.} The cognitive and background aspects are aggregated into the \emph{Professional Competency} track, and the social aspects form the \emph{Communication Competency} track. The overall score is the mean of these two tracks. A final CV-informed step adapts the recommendations to the candidate's experience and target role.

\paragraph{Theory grounding and traceability.} The KSA framework specifies the job-relevant competencies represented by the professional track \citep{peterson2001onet}, while STAR structures the evidence expected in behavioral responses \citep{janz1982behavior}. For example, an omitted Result can be converted into a targeted recommendation to state the outcome and its significance. Each competency score can be traced through its contributing aspects to the behavior-level features and modalities that produced it, linking theoretical criteria to inspectable evidence. Appendix~\ref{app:theory} details how KSA, STAR, and the Big Five enter the assessment pipeline.

\section{Implementation and Deployment}
\label{sec:impl}

\paragraph{Real-time pipeline.} The digital human interviewer is driven by lip-sync generation \citep{prajwal2020wav2lip} and streamed to the browser over WebRTC, with streaming speech recognition and synthesis supporting the spoken interaction. LLM-based agents manage question generation and follow-up reasoning, a VLM analyzes non-verbal behavior \citep{bai2023qwenvl}, and a speech service provides pronunciation analysis. The platform runs as four cooperating services behind HTTPS. A session pool caps concurrent digital human sessions at five and queues additional requests. The four evaluators run concurrently for each response through a thread pool. Model versions, service layout, and stack details are provided in Appendix~\ref{app:stack}.

\paragraph{Demo interaction.} A visitor selects an interviewer persona and either uploads a CV or uses a bundled sample CV and JD. After a brief spoken interview with three or four questions, the visitor can inspect the report from the two competency tracks through the assessment aspects to individual behavior-level features and recommendations. A prerecorded end-to-end session and exported PDF report provide a network-independent demonstration fallback.

\section{System Usage Analysis}
\label{sec:deploy}

\begin{table}[t]\centering\small
\begin{tabular}{@{}p{0.67\linewidth}r@{}}
\toprule
\textbf{Statistic} & \textbf{Snapshot} \\
\midrule
Accounts & 101 \\
Returning accounts & 62 \\
Sessions & 1{,}564 \\
Generated questions & 7{,}665 \\
Completed sessions & 217 \\
Five-stage question sets & 1{,}422 \\
Distinct position titles & 83 \\
Answered main-question records & 1{,}274 \\
Recorded follow-up turns & 342 \\
Pooled scored response records & 1{,}425 \\
Audio/video files (wav/webm) & 1{,}744 / 1{,}744 \\
\bottomrule
\end{tabular}
\caption{All-account usage snapshot, including internal and test activity.}
\label{tab:deploy}
\end{table}

\paragraph{Usage scale and workflow coverage.} PolyInterview runs on the university network. The snapshot includes 101 account directories and 1{,}564 sessions, including test, internal, placeholder, and team accounts, so these totals characterize platform activity rather than verified external candidates. It contains 7{,}665 generated questions across 83 position titles, and 1{,}422 session question sets (90.9\%) cover all five planned categories: self-introduction, behavioral, skill-QA, scenario, and candidate questions (Table~\ref{tab:deploy}). The snapshot also contains 1{,}744 WAV and 1{,}744 WebM recordings, demonstrating capture of both spoken and visual interview behavior.

\begin{figure}[t]
  \centering
  \includegraphics[width=\linewidth]{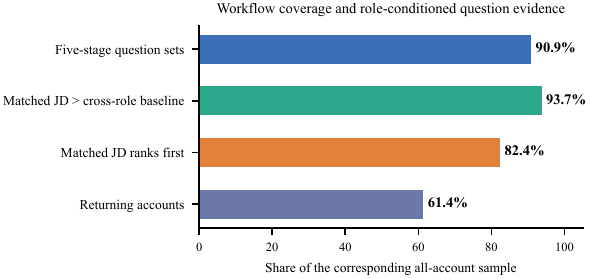}
\caption{Workflow coverage, role alignment, and returning-account evidence.}
  \label{fig:deployment-evidence}
\end{figure}

\paragraph{Role-conditioned question generation.} We compare each session's question set with its own JD and JDs from other position groups. The median similarity is 0.055 for the matched JD and 0.0058 across roles, about a 9.5-fold ratio. The matched JD exceeds the cross-role baseline in 93.7\% of sessions and ranks first in 82.4\% (Figure~\ref{fig:deployment-evidence}). This lexical test does not replace expert judgment, but shows that the generated questions differ systematically by target role.

\paragraph{Adaptive interaction and multimodal assessment coverage.} The logs contain 1{,}274 answered main questions, including 208 with follow-ups and 342 follow-up turns. The assessment pool contains 1{,}425 scored responses because it includes main and follow-up answers. These counts and the recordings above show that personalization, multi-turn interaction, and multimodal assessment operate in the deployed system.

\section{Human Expert Evaluation of System Outputs}
\label{sec:human-expert}

To evaluate the quality of PolyInterview's system outputs, we conducted a human expert study. Ten experts with backgrounds including HR, language assessment, and communication reviewed the same 12 de-identified, text-visible artifacts: four role-conditioned question plans, four response and follow-up pairs, and four evidence-linked feedback reports. They scored three module-specific criteria from 1 to 5, yielding 120 artifact reviews and 360 criterion ratings.

\begin{center}
\small
\setlength{\tabcolsep}{3.5pt}
\begin{tabular}{@{}lrrrr@{}}
\toprule
\textbf{Module} & \textbf{$n$} & \textbf{Mean} & \textbf{Expert SD} & \textbf{$\pm1$ agree} \\
\midrule
Question plans & 4 & 4.62 & 0.15 & 100.0\% \\
Follow-ups & 4 & 3.68 & 0.18 & 98.9\% \\
Feedback reports & 4 & 3.74 & 0.37 & 91.3\% \\
\bottomrule
\end{tabular}
\captionsetup{hypcap=false}
\captionof{table}{Scores from the evaluation by ten human experts on a scale from 1 to 5.}
\label{tab:llm-panel}
\end{center}

Experts rated question plans most strongly (4.62/5), with high role relevance (4.80) and stage coverage (4.83). Follow-ups were relevant (4.65) but weaker in answer dependence (3.28) and diagnostic depth (3.13). Feedback recommendations were actionable (4.70), while polished-response faithfulness was lower (2.75) because some examples introduced unsupported context. These results identify strengths and targets for improving follow-up reasoning and response faithfulness.

\section{Conclusion}
PolyInterview combines questions tailored to the JD and CV with answer-aware digital human interaction and multimodal assessment grounded in KSA and STAR. Its hierarchy connects 13 behavior-level features to 10 assessment aspects and two competency tracks, tracing guidance to multimodal observations. Across 101 accounts, 1{,}564 sessions generated 7{,}665 questions, with 90.9\% of question sets covering all five stages and 82.4\% ranking the matched JD first. Ten experts found strong question plan quality and actionable recommendations, while identifying follow-up depth and response faithfulness as improvement targets.

\clearpage
\section*{Limitations}
The usage analysis is observational and the all-account totals intentionally include internal, test, placeholder, and team activity. They therefore measure platform activity rather than unique external candidates. The role-alignment analysis measures lexical correspondence and still requires expert validation of relevance and quality. The human-expert study evaluates representative artifacts rather than longitudinal outcomes. Because the deployment is recent, we do not yet know whether repeated practice translates into later interview success or employment outcomes. We therefore plan opt-in follow-up surveys with deployed users to track perceived skill gains and subsequent outcomes. Controlled pre/post studies would still be needed to isolate causal effects. Artifact-level expert ratings also do not establish score-level criterion validity, which requires blinded comparison against independent career-service ratings. The interface panels in Figure~\ref{fig:journey} are illustrative walkthroughs from different demo sessions rather than evidence of typical performance.

\section*{Ethics Statement}
PolyInterview is an advisory practice tool for candidates, explicitly not a hiring or gatekeeping system, and its scores are explainable and contestable. Automated assessment of non-verbal behavior has a documented history of fairness concerns. We therefore keep visual assessment advisory, expose supporting evidence for every score, and avoid any pass/fail decision. The platform processes potentially sensitive data, including CVs, response transcripts, voice recordings, and interview video. It is deployed through campus servers, with encrypted transmission and storage, strict access controls, and per-user data isolation. Raw data are retained only to provide the user-facing service and are not retained for secondary use. Users may delete their interview records and associated files at any time, after which they are removed from the server. Neither PolyInterview nor its external model providers use raw user data to train or fine-tune models, and provider calls operate under no-training terms. Any research use is de-identified and IRB-approved, consistent with GDPR and Hong Kong's PDPO. The analysis in this paper uses de-identified session metadata only.

\bibliography{references}

\clearpage
\appendix
\section{Detailed Related Work}
\label{app:related}

\paragraph{Commercial interview tools.} Google Interview Warmup, Yoodli, Big Interview, HireVue, and Interviewing.io support question rehearsal, selected delivery metrics, response examples, or human coaching. Newer LLM-based products additionally offer resume-tailored sessions and per-question feedback. However, as Table~\ref{tab:compare} summarizes, these tools generally do not combine answer-aware questioning, comprehensive multimodal assessment, and evidence-linked feedback within one multi-stage practice workflow.

\paragraph{Virtual interview agents.} Earlier HCI systems addressed complementary parts of the problem. MACH provides video-aligned coaching on smiles, prosody, and filler words \citep{hoque2013mach}. TARDIS/Gloria senses social cues in real time but follows a predetermined scenario \citep{anderson2013tardis,baur2013job,langer2016dear}. VR-JIT explains prescripted response options rather than free speech \citep{smith2014virtual}. ERICA later generated follow-ups from answer quality but did not provide post-interview feedback \citep{inoue2020jobinterviewer}. Thus, these systems offer multimodal delivery analysis or explainable content feedback, but not both alongside answer-aware probing.

\paragraph{LLM-based mock interviews.} LLMs enable flexible question generation \citep{li2023ezinterviewer}, deployed end-to-end interviewing \citep{wang2023interviewbot}, simulations that match people to jobs \citep{sun2025mockllm}, talking-head interaction \citep{nguyen2025siminterview}, and transcript-grounded feedback \citep{daryanto2025conversate}. Systems that provide assessment still tend to operate on text alone or cover only selected communication channels. PolyInterview instead connects adaptive dialogue to a shared rubric spanning textual, vocal, and non-verbal evidence.

\paragraph{Multimodal and model-based assessment.} Automated video interview models predict interview constructs from multimodal features \citep{naim2018automated,hemamou2019hirenet}, but reliability and validity vary across constructs and individual scores can be difficult to inspect \citep{hickman2022automated}. Foundation-model evaluators carry additional judgment biases \citep{zheng2023judging}. PolyInterview therefore keeps visual assessment advisory and exposes the intermediate features, aspect mappings, and supporting evidence behind its recommendations.

\section{Technology Stack and Service Details}
\label{app:stack}
The frontend is a Vue single-page application, a Flask API orchestrates interview flow and assessment, and a separate Node/Express service handles authentication. Qwen-Plus/Qwen-Flash provide text scoring, Qwen3-VL-Plus analyzes non-verbal behavior \citep{bai2023qwenvl}, and Qwen3-ASR supports streaming recognition. Edge-TTS provides synthesis and Azure Speech provides pronunciation scoring. The digital human uses LiveTalking with Wav2Lip \citep{prajwal2020wav2lip} and streams over WebRTC. Four services (frontend, API, authentication, and digital human) run behind HTTPS. Per-answer WebM recordings are converted to MP4 and WAV for modality-specific analysis.

\section{Adaptive Interviewing Details}
\label{app:adaptive}

Personalization operates both before and during the interview. The planner constructs the initial question sequence from the setup inputs, while the interviewer adapts subsequent dialogue to the candidate's responses.

\begin{figure}[t]
  \centering
  \includegraphics[width=\linewidth]{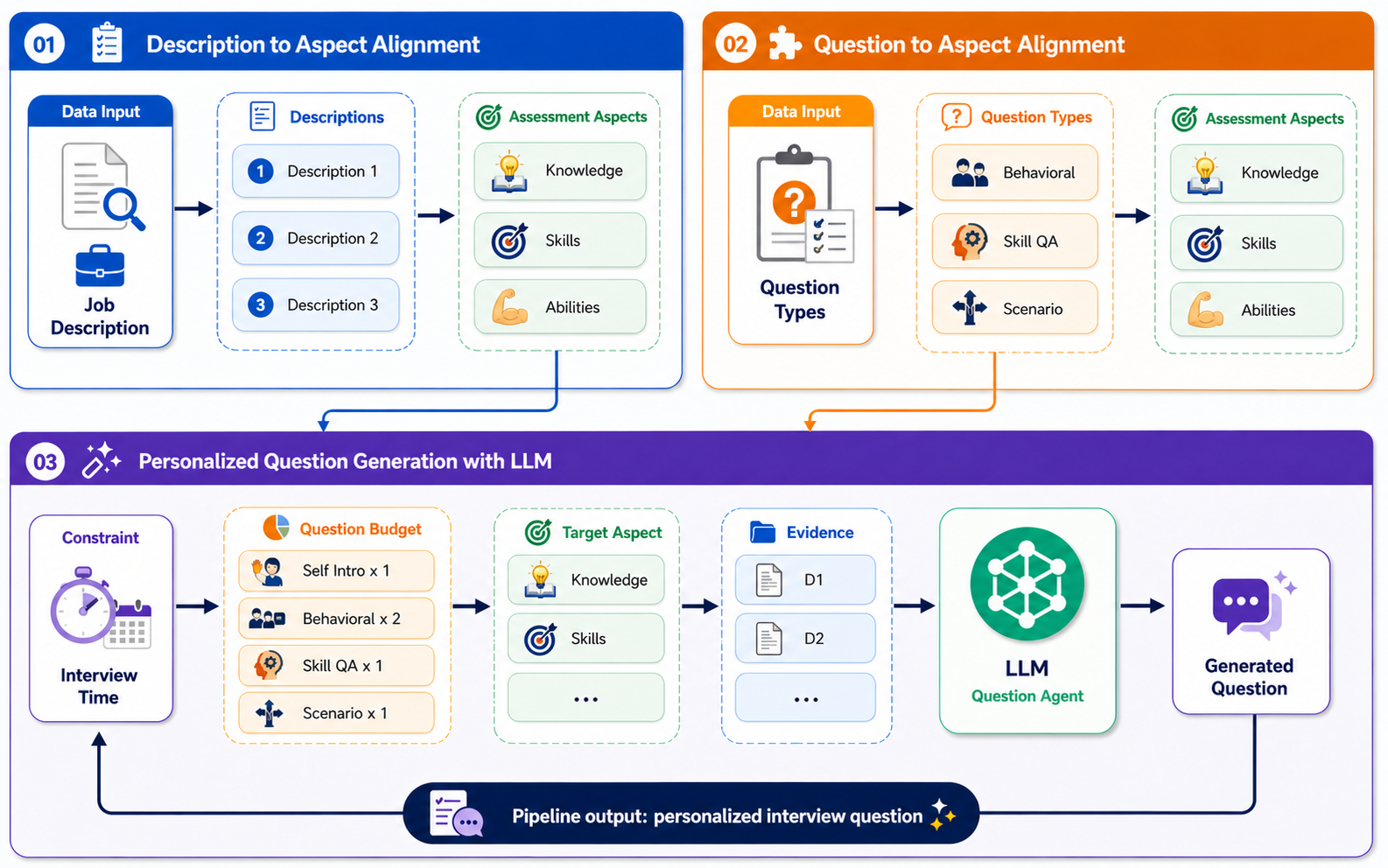}
\caption{Personalized question generation through description-to-aspect and question-to-aspect alignment.}
  \label{fig:qgen}
\end{figure}

\paragraph{Question planning.} The planner receives the target company and position, JD, CV, interviewer persona, and session duration (Figure~\ref{fig:qgen}). It aligns role requirements and question categories with KSA-based assessment aspects, allocates the duration-dependent budget across self-introduction, behavioral, skill-QA, scenario, and candidate-question stages, and grounds each question in relevant JD or CV evidence. This connects personalization to the same competencies used in the subsequent assessment.

\begin{figure}[t]
  \centering
  \includegraphics[width=\linewidth]{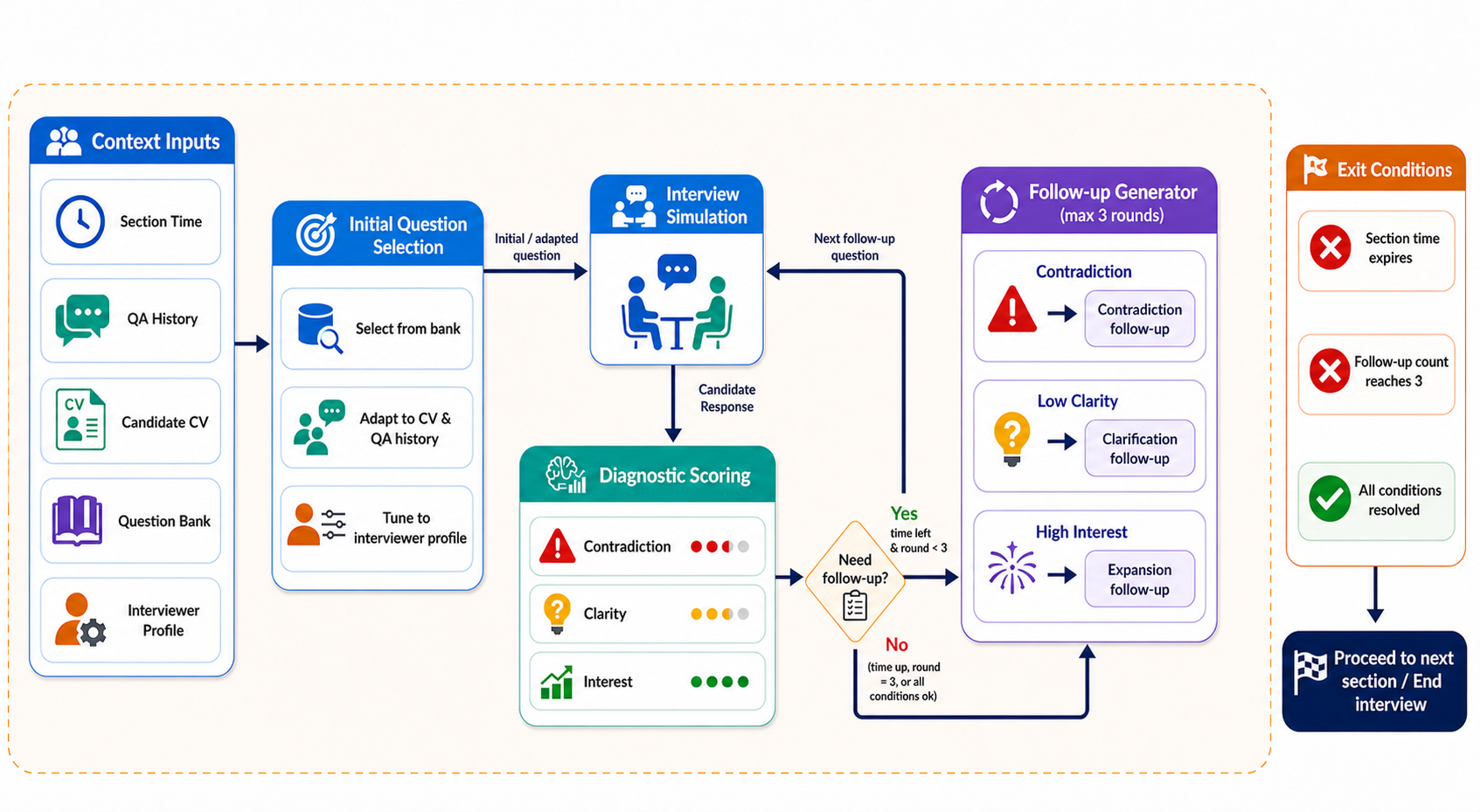}
\caption{Answer-aware follow-up generation using diagnostic scoring, bounded follow-up rounds, and explicit exit conditions.}
  \label{fig:followup}
\end{figure}

\paragraph{Real-time follow-ups.} The interviewer maintains the dialogue context and evaluates each response for contradiction, clarity, and interest (Figure~\ref{fig:followup}). It can request resolution of a contradiction, clarification of an underspecified response, or elaboration on a point of interest. The interview proceeds once the response is sufficiently resolved or a section-time or follow-up limit is reached. Basic, Intermediate, and Advanced personas allow at most one, two, and three follow-ups, respectively, adapting both question content and probing depth.

\section{Assessment Frameworks: KSA, STAR, and the Big Five}
\label{app:theory}
The three frameworks enter the pipeline at different points rather than serving only as background motivation.

\paragraph{KSA: competency specification.} Knowledge, Skills, and Abilities provides the job-analysis vocabulary used to organize the cognitive and background aspect families (Table~\ref{tab:mapping}) and the Professional Competency track. Aligning planned questions with these aspects connects role requirements to the criteria later used for assessment.

\paragraph{STAR: evidence structure.} Situation, Task, Action, Result structures the evidence expected in behavioral responses \citep{janz1982behavior}. PolyInterview uses it to assess whether a candidate explains the context, goal, personal action, and outcome. A missing element can therefore produce a specific recommendation rather than a generic request for more detail.

\paragraph{Big Five: grounding social constructs.} The five-factor model \citep{mccrae1987validation} informs the selection of communication, interpersonal skills, leadership, and persuasiveness as socially relevant constructs. PolyInterview does not infer or report personality traits. The framework only prevents this part of the rubric from relying on ad-hoc labels.

\section{Mapping Assessment Aspects to Behavior-Level Features and Question Gating}
\label{app:mapping}
Each assessment aspect applies a 70/30 weighting to its primary and secondary feature groups (Table~\ref{tab:mapping}). Question gating activates only aspects that a category can reasonably elicit. Self-introduction targets Communication Skills, Education \& Training, and Experience. Behavioral questions target Interpersonal Skills, Applied Mental Skills, and Leadership. Skill-QA targets Job Knowledge, General Intelligence, and Education \& Training. Scenario questions target Applied Mental Skills, Creativity, and Persuasiveness. Candidate questions are not scored. This mapping traces a competency score through its assessment aspects to the underlying textual, vocal, and visual evidence.

\begin{table}[h]
\centering\footnotesize
\setlength{\tabcolsep}{2pt}
\begin{tabular}{@{}>{\raggedright\arraybackslash}p{2.05cm}>{\raggedright\arraybackslash}p{2.6cm}>{\raggedright\arraybackslash}p{2.5cm}@{}}
\toprule
\textbf{Aspect} & \textbf{Primary (70\%)} & \textbf{Secondary (30\%)} \\
\midrule
\multicolumn{3}{@{}l}{\emph{Cognitive}} \\
Gen.\ Intelligence & ConcAcc, Logic & Coher, Clarity \\
Applied Mental & Logic, ConcAcc & Clarity, Coher \\
Creativity & Logic, Word & ConcAcc, Coher \\
\midrule
\multicolumn{3}{@{}l}{\emph{Background}} \\
Job Knowledge & ConcAcc, Term & Logic \\
Education \& Train. & ConcAcc, Term & Clarity \\
Experience & ConcAcc, Logic, Word & Term \\
\midrule
\multicolumn{3}{@{}l}{\emph{Social}} \\
Communication & Clarity, Coher, Word & Eye, Face, Pron \\
Interpersonal & Eye, Face, Coher & Posture, Gest, Clarity \\
Leadership & Logic, Posture, Eye & Clarity, Face, Gest \\
Persuasiveness & Clarity, Coher, Word & Eye, Face, Prosody \\
\bottomrule
\end{tabular}
\caption{Primary and secondary behavior-level features used to compute each assessment aspect.}
\label{tab:mapping}
\end{table}

\paragraph{Reading the mapping.} ConcAcc, Term, and Logic denote conceptual accuracy, terminology use, and problem-solving logic. Clarity, Coher, and Word denote structural clarity, coherence, and word usage. Eye, Face, Posture, and Gest are visual features, while Pron and Prosody describe vocal delivery. Primary and secondary groups indicate relative contribution.

Gating selects only assessment aspects that a question can elicit. Inactive aspects remain unscored. The report traces each competency track through its aspects and behavior-level features to the supporting evidence, attributing lower scores to specific content, delivery, or non-verbal signals.

\end{document}